\documentclass[10pt,twocolumn,letterpaper]{article}

\usepackage{arxiv}
\usepackage{times}
\usepackage{epsfig}
\usepackage{graphicx}
\usepackage{amsmath}
\usepackage{amssymb}
\usepackage{multirow}
\usepackage{diagbox}
\usepackage{caption}
\usepackage{subfigure}
\usepackage{color}

\usepackage[breaklinks=true,bookmarks=false]{hyperref}

\cvprfinalcopy 

\begin{document}
	
	\title{Representation based and Attention augmented Meta learning}
	
	\author{Yunxiao Qin
	\footnotemark[1]\\
	{\tt\small qyxqyx@mail.nwpu.edu.cn}
	\and
	Chenxu Zhao
	\footnotemark[1]\\
	{\tt\small zhaochenxu1@jd.com}
	\and
	Zezheng Wang
	\footnotemark[1]\\
	{\tt\small wangzezheng1@jd.com}
	\and
	Junliang Xing\\
	{\tt\small jlxing@nlpr.ia.ac.cn}
	\and
	Jun Wan\\
	{\tt\small jun.wan@nlpr.ia.ac.cn}
	\and
	Zhen Lei\\
	{\tt\small zlei@nlpr.ia.ac.cn}
}
	
	\author{Yunxiao Qin$^{1,2*}$, Chenxu Zhao$^{3*}$, Zezheng Wang$^{2*}$, Junliang Xing$^4$, Jun Wan$^4$  Zhen Lei$^4$\\
		$^1$Northwestern Polytechnical University of China \\
		$^2$JD Finance, $^3$JD AI Research, $^4$Chinese Academy of Sciences \\
		qyxqyx@mail.nwpu.edu.cn, \{qinyunxiao, zhaochenxu1, wangzezheng1\}@jd.com\\
		 \{jlxing, jun.wan, zlei\}@nlpr.ia.ac.cn
	}

	\maketitle
	\renewcommand{\thefootnote}{\fnsymbol{footnote}}
	\footnotetext[1]{ denotes equal contribution.}
	\begin{abstract}
		Deep learning based computer vision fails to work when labeled images are scarce. Recently, Meta learning algorithm has been confirmed as a promising way to improve the ability of learning from few images for computer vision. However, previous Meta learning approaches expose problems: 
		1) they ignored the importance of attention mechanism for the Meta learner; 
		2) they didn't give the Meta learner the ability of well using the past knowledge which can help to express images into high representations, resulting in that the Meta learner has to solve few shot learning task directly from the original high dimensional RGB images. 
		In this paper, we argue that the attention mechanism and the past knowledge are crucial for the Meta learner, and the Meta learner should be trained on high representations of the RGB images instead of directly on the original ones. Based on these arguments, we propose two methods: Attention augmented Meta Learning (AML) and Representation based and Attention augmented Meta Learning(RAML). The method AML aims to improve the Meta learner's attention ability by explicitly embedding an attention model into its network. The method RAML aims to give the Meta learner the ability of leveraging the past learned knowledge to reduce the dimension of the original input data by expressing it into high representations, and help the Meta learner to perform well. Extensive experiments demonstrate the effectiveness of the proposed models, with state-of-the-art few shot learning performances on several few shot learning benchmarks. The source code of our proposed methods will be released soon to facilitate further studies on those aforementioned problem. 

	\end{abstract}
	
	\section{Introduction}
	Deep learning based computer vision system has recently achieved great success and has shown its outstanding performance in many applications, such as object classification\cite{image-classification1,image-classification2,image-classification3}, semantic segmantation\cite{seg1,seg2}, face recongnition\cite{face1,face2} etc. 
	
	However, the deep learning based computer vision is still struggling when labeled data is scarce\cite{match-network, zhuang}. 
	As a comparison, human vision system is so smart that human can efficiently learn new object with great performance from few images. Obviously, current computer vision still lag behind human vision greatly. 
	
	Recently, several Meta learning approaches\cite{Learning-a-synaptic,On-the-optimiazation,An-alternative-to,MAML,Meta-SGD,miniimagenet,Reptile,SNAIL}have improved the ability of learning from few data for the computer vision system. 
	Different from the ordinary deep learning approaches which train the network on a distribution of data, the Meta learning approach trains a Meta learner on a distribution of tasks instead of data, so that the Meta learner can generalize well on tasks never seen before which are sampled from this distribution. 
	
\begin{figure}[t]
		\centering    
		\subfigure[Meta learner A] 
		{
			\includegraphics[scale=0.45]{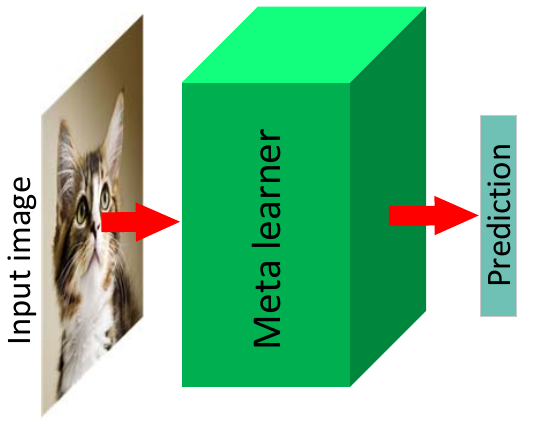}  
			\label{fig:a} 
		}
		\subfigure[Meta learner B] 
		{
			\includegraphics[scale=0.45]{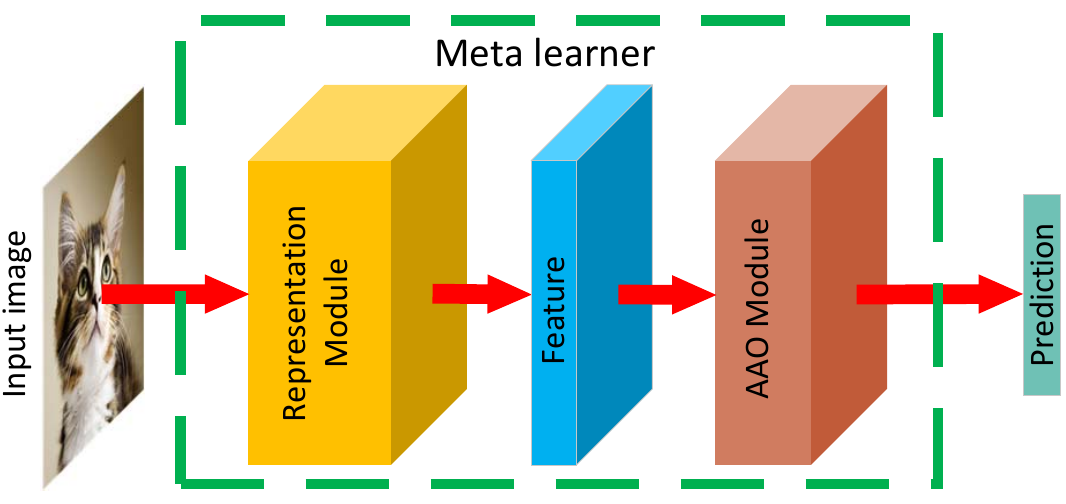}
			\label{fig:b}
		}
		\caption{(a) The Meta learner which is trained to accurately adjust its total network according to the original input data. (b) The Meta learner which is separated into representation module and AAO module. The representation module is a module that well using its past knowledge to reduce the dimension of the input data by expressing the data into high representations (feature), and the AAO module is trained to quickly adjust itself based on the high representations.} %
		\label{fig:struture of previous Meta learner and the RAML's}  
	\end{figure}

	Even with the above improvement, previous Meta learner still did not shorten the distance with human vision greatly, which is mainly because:
	1) given a specific task with few data, and no attention mechanism\cite{soft_attention1,soft_attention2,soft_attention3,hard_attention,attention-is-all-you-need,human_attention1,human_attention2}, these Meta learners are not capable of paying attention to the most distinguishable features of the input images to solve this task; 
	2) these Meta learners can not well use the past learned knowledge to express the input images into high representations accurately, and they have to update themselves quickly according to the few original high dimensional input RGB images, which is shown as Figure.\ref{fig:a}.

	In this paper, we propose our idea that the attention mechanism is important for the Meta learner, and the Meta learner should take full advantage of the past learned knowledge to accurately express the input data into high representations with low dimension, and the Meta learner should update itself according to the high representations instead of the original input data, which is shown as Figure.\ref{fig:b}. 
	
\begin{figure}
		\centering
		\includegraphics[width=0.4\textwidth]{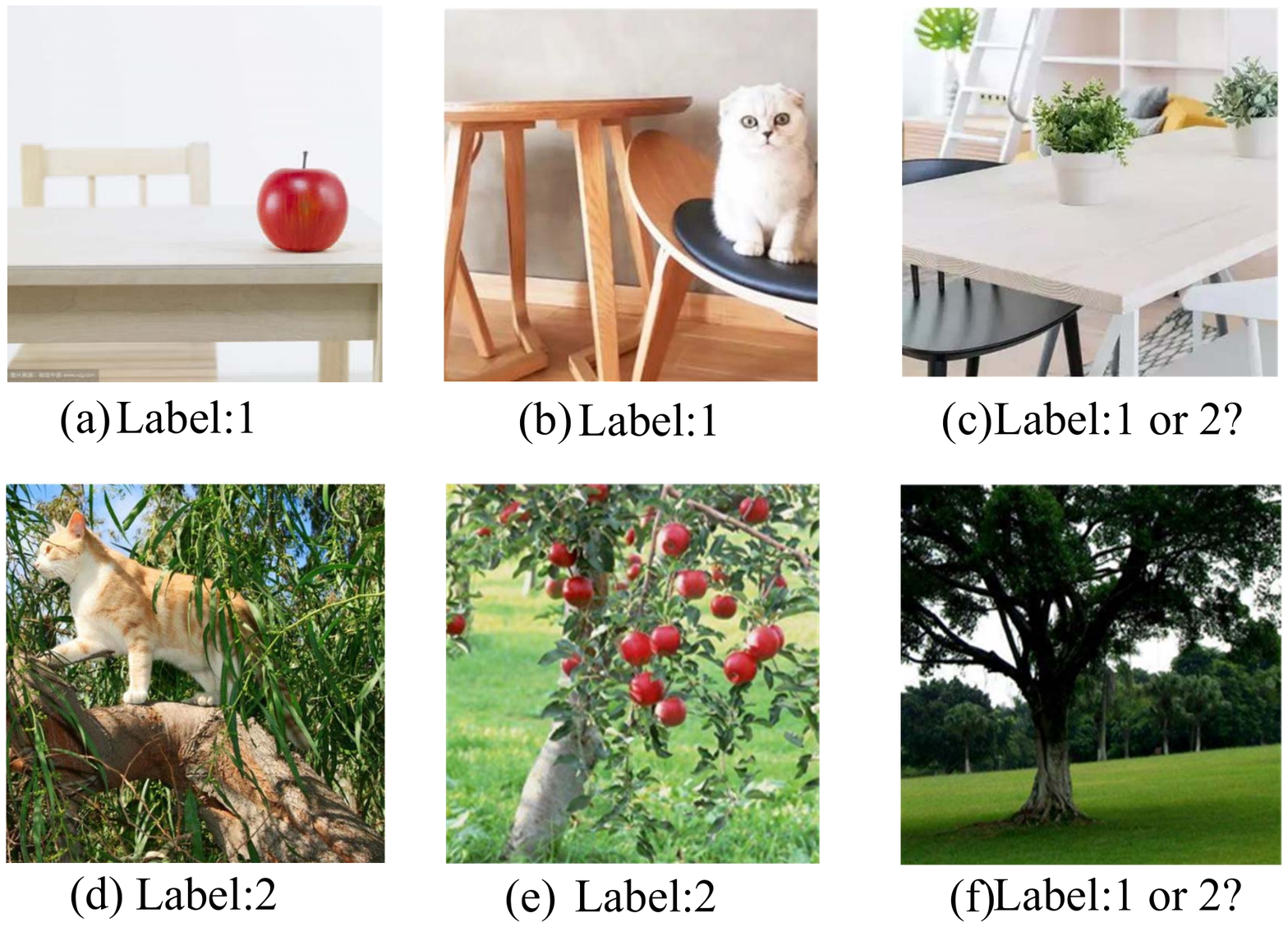}
		\caption{A simple few shot image classification task.}
		\label{fig:few shot example}
	\end{figure}

	To better understand the motivation of our idea, it is necessary to take an example which shows in Figure.\ref{fig:few shot example} to briefly analyze the few shot learning process of human intelligence. 
	In Figure.\ref{fig:few shot example}, there are six images belonging to two image classes, 4 labeled images are training data, and 2 unlabeled images are testing data. Now, we are required to category the two unlabeled images, image (c) is probably labeled as 1, due to containing table. This feature is not only the same feature with the other images labeled as 1, but also different from the images labeled as 2. Similarly, image (f) is probably labeled as 2, due to containing tree. 
	
	In this example, these images have several features, such as plant, animal, tree, table etc. However, only the feature of containing tree or table is useful for us to category these two images correctly, and we quickly pay attention to the key feature and neglect the others, manifesting the important role of attention mechanism. 
	
	It should be noted that it is easy for us to get these meaningful features of these images, because we have learned the knowledge about the world before the few shot learning task, and we well use our knowledge to express these images into high representations exactly. 
	Meanwhile, our expression about these images are constant, but we have to quickly adjust our attention and decision logical to fit this task. 
	
	Therefore, we can see two main modules in the few shot learning process: a representation module that is to utilize the past knowledge to express images into meaningful high representations, and an attention based decision logical module that is to update quickly to fit the new few shot learning task. 
	
	In the above example, it is obvious that we didn't do the few shot learning task directly based on the high dimensional RGB images. Instead, we firstly express each image into high representations with lower dimension space, and secondly reduce the representation's dimension by quickly changing our attention to select the key feature of the high representations. 
	Based on the feature which we select, it is exceedingly easy for us to adjust our decision logic to fit the few shot learning task. 
	As a comparison, previous Meta learning approaches trains the Meta learner to do a much harder work, that is, the Meta learner is forced to quickly adjust its total network to fit the few shot learning task based on the original high dimensional input images. 
	
	Inspired by the few shot learning process of human vision system, and to help the Meta learner to do a better work, we propose two methods:  
	1) Improving Meta learner's ability of learning from few data by explicitly embedding attention mechanism into it. We call this method as Attention augmented Meta learning(AML). This method makes the Meta learner be capable of paying more attention on the key feature. 
	2) Easy the Meta learner's work by separating its total network into two modules: representation module and Attention Augmented Output(AAO) module. The architecture of the network is shown in Figure.\ref{fig:b}, and these two modules are trained separately in two stages. The representation module should be pre-trained by supervised learning to learn enough knowledge, and it is equivalent to the same module of human intelligence.

	In the Meta learning stage, the Meta learner will be trained to quickly adjust the AAO module based on the high representations that the pre-trained representation module provides. Obviously, the AAO module plays the same role as the attention based decision logical module of human vision system. We call this method as Representation module based and Attention augmented Meta learning(RAML).
	
	The main contributions of our work are:
	\begin{itemize}
		\vspace{-7pt}
		\item Analyze the problems of previous Meta learning approaches, and propose our viewpoint that both the attention mechanism and the past knowledge are crucial for the Meta learner, and the Meta learner should be trained on high representations of the input data instead of the original data; 
		\vspace{-7pt}
		\item Based on our viewpoints, we design two methods: Attention augmented Meta learning(AML) and Attention augmented Meta learning(RAML).
		\vspace{-7pt}
		\item By lots of experiments, we attained state-of-the-art performance on several few shot learning benchmarks both with methods AML and RAML, showing the rationality of our viewpoint and methods. 

	\end{itemize}
	
	\section{Related Work}
	\subsection{Meta learning}
	
	A N-way, K-shot learning task means there is a support set and a query set for the Meta learner. The support set contains K examples for each of the N classes, and the query set contains L examples for each of the N classes. Meta learning has been shown as a promising way to solve the few shot learning problem, and most of the Meta learning approaches train the Meta learner on the N-way, K-shot learning tasks in the following way: firstly, the Meta learner is required to inner-update itself according to the support set; secondly, the Meta learner exams the effect of the inner-update operation by calculating its loss on the query set,  thirdly, by minimizing the loss on the query set, the Meta learner is forced to learn a great weight initializer (the initialized weight is easily inner-updated by simplely gradient descent to perform well on the query set\cite{MAML}), or a skillful weight updater (accurately inner-updating the Meta learner's weight \cite{miniimagenet}), or both\cite{Meta-SGD}, or to memorize the information of the support set and to perform well on the query set based on the memory\cite{SNAIL}.
	
	Reptile\cite{Reptile} which trains the Meta learner only with first order gradient also force the Meta learner to learn the weight initializer. 
	In our paper, we use the Meta-SGD approach\cite{Meta-SGD} as our basic Meta learning approach for its excellent performance and feasibility.
	\vspace{-3pt}
	\subsection{Metric Learning}
	\vspace{-5pt}
	Some researchers have tried to solve the few shot learning problem by metric learning. The principal of these approaches is straight-forward, that is, to train a non-linear mapping function that represents images in an embedding space. After this training, the embedding of images belonging to different classes are easy to be distinguished by simple nearest neighbor or linear classifiers.
	Matching network\cite{match-network} trains the mapping function using a neural network, and categorizes the test images to the class where images have the most similar embedding with them, and the similarity is measured by the Cosine distance between embedding. 
	Prototypical Network\cite{prototypical} is an approach similar to Matching network, whereas the similarity between embedding is measured by Euclidean distance. 
	Compared with Meta learning approaches, the disadvantage of the above kinds of metric learning based approaches is obvious: they are not easily applicable to other domains, such as reinforcement learning\cite{DRL1, DRL2} and regression.
	\section{Algorithm}
	
	\subsection{Problem of learning from few data}
	Learning from few data is extremely difficult for the deep learning based computer vision system. This is mainly because there are too much weight in the deep neural network, and  the input data is represented in a large dimension space, usually tens or hundreds of thousands dimension space is required. 
	
	For the image classification task, it is difficult for few images, in such a large dimension space of one category, to reflect the characteristic of this category accurately. 
	However, human vision system can get the characteristic of a new category from few images 

	by firstly expressing them into high representations and secondly paying attention to the key features of the high representations, and both help human to understand the characteristic of the category with few images.
	
	Previous Meta learning approaches help computer vision system to learn from few data a lot. However, they train the Meta learner to quickly adjust its network to fit the few shot learning task directly on the few original high dimensional input images, and ignore the importance of attention mechanism and the past knowledge, which can help the Meta learner to perform well.
	
	In this paper, to counter the problem the previous Meta learning approaches expose, we propose two methods: Attention augmented Meta Learning(AML) and Representaion based and Attention augmented Meta Learning(RAML). 
	
	\begin{figure}
		\centering
		\includegraphics[width=0.49\textwidth]{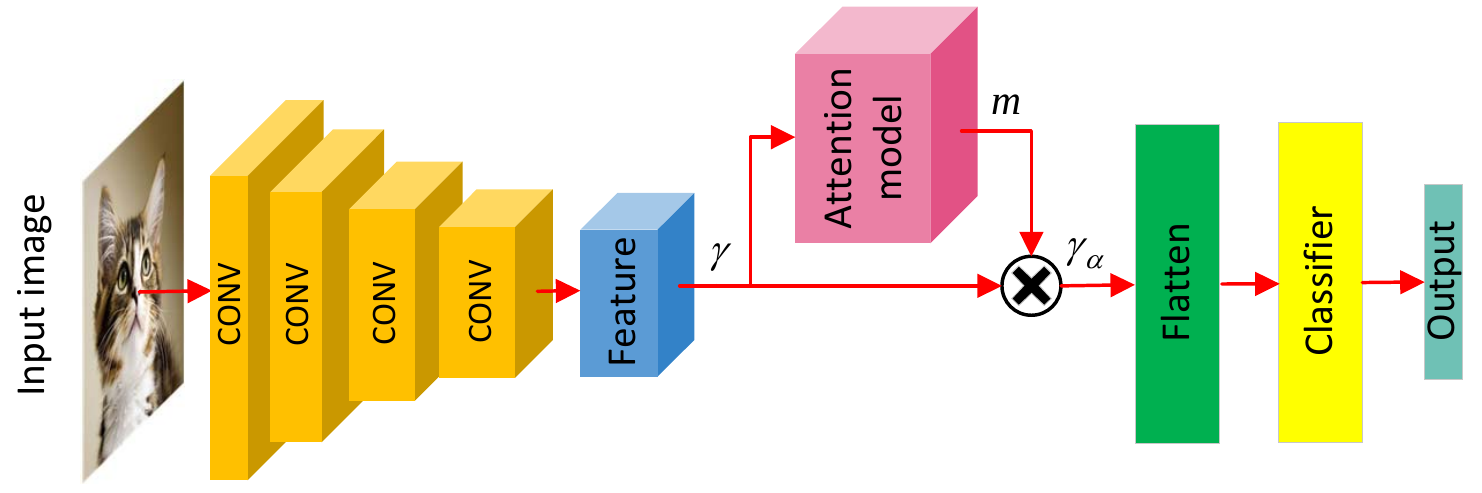}
		\caption{The network structure of the Meta learner in method AML. There is an attention model inserted explicitly in the Meta learner's network}
		\label{fig:network structure of AML}
	\end{figure}

	\begin{figure*}[t]
		\centering    
		\subfigure[] 
		{
			\includegraphics[scale=0.6]{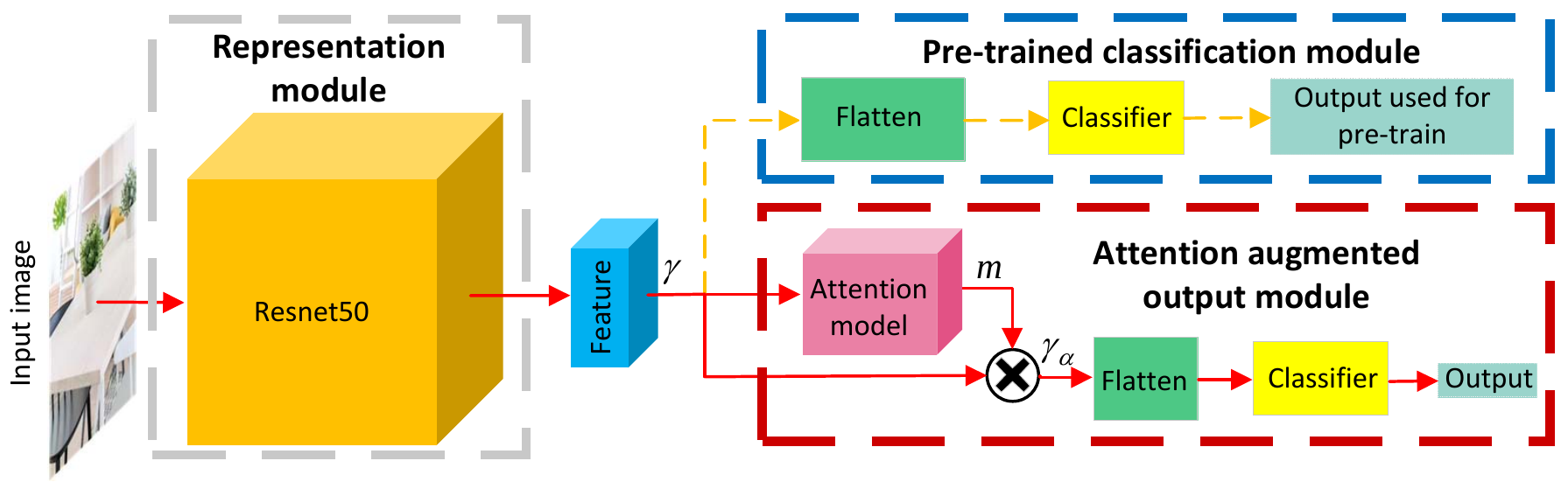}  
			\label{fig:4a} 
		}
		\subfigure[] 
		{
			\includegraphics[scale=0.6]{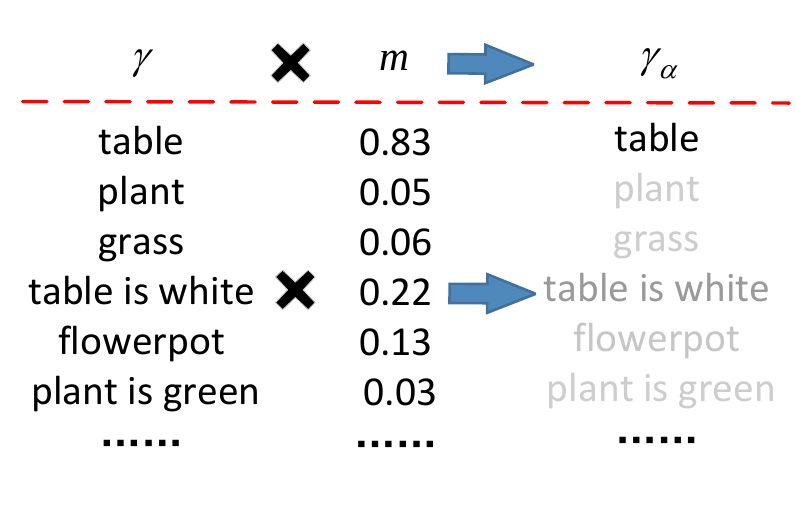}
			\label{fig:4b}
		}
		\caption{(a) The network structure used in method RAML. The Meta learner is composed of representation module and the AAO (attention augmested output) module. The PC (pre-trained classification) module is used to pre-train the representation module.
		(b) An example that interpreting the principal of soft attention mechanism for few shot learning.} 
		\label{fig:network structure of RAML}  
	\end{figure*}

	\vspace{-1pt}
	\subsection{AML}
	\vspace{-3pt}
	Method AML aims to improve the Meta learner's attention ability by explicitly embedding an attention model into its network, and the attention model will help the Meta learner to pay attention on the key features. 
	
	The network architecture of the Meta learner is shown in Figure.\ref{fig:network structure of AML}. An attention model is inserted explicitly, and the forward calculation is shown as Eq.~\ref{eq:AML}. The feature $\gamma$ which the CNN outputs is firstly fed into the attention model, and the final feature $\gamma_\alpha$ is the channel-wise multiplication between the attention mask \emph{m} and the original feature $\gamma$, and the classifier output the final prediction \emph{p}. 
	\begin{equation}
	\begin{split}
	\gamma &= F_{CNN}(x; \theta_{cnn}), \\
	m &= F_{Att}(\gamma; \theta_{att}), \\
	\gamma_\alpha &= F_{channel}(\gamma, m), \\
	p &= F_{Cla}(\gamma_\alpha; \theta_{cla}),
	\label{eq:AML}
	\end{split}
	\end{equation}
	
	Where \emph{$F_{CNN}$}, \emph{$F_{Att}$}, \emph{$F_{channel}$}, \emph{$F_{Cla}$} are the functions of the CNN, attention model, channel-wise multiplication and the classification layer respectively. Moreover, x is the input data, $\theta_{cnn}$ is the weight of the CNN,  $\theta_{att}$ is the weight of the attention model, and \emph{p} is the final prediction of the classification layer with the weight $\theta_{cla}$. 
	
	By embedding the attention model explicitly, the Meta learner will be more capable to adjust its attention to the more useful feature in the few shot learning task, and will help the classifier to do a better few shot learning work. 
	Corresponding experiments show the positive effect of attention mechanism, and we analyze the feature distribution of $\gamma$ and $\gamma_\alpha$ in the feature analysis section. It is clear that $\gamma_\alpha$ is more distinguishable than $\gamma$.

	\subsection{RAML}
	Method RAML aims to give the Meta learner the ability of both well using the past learned knowledge and the attention mechanism. To achieve that, we seperate the Meta learner's network into two modules: representation module and AAO module. 
	The representation module is used to learn knowledge on the other dataset by supervised learning, e.g. MiniImagenet-900 dataset(a dataset we organized to pre-train the representation module, and the detail about it will be introduced in the experiment section). 
	The AAO module is the module which also embeds an attention model and can be adjusted efficiently to fit the new few shot learning task by the Meta learner.

	The network structure of Meta learner is shown in Figure.\ref{fig:4a}. 
	The Pre-trained Classification(PC) module does not belong to the Meta learner, and it is only used to pre-train the representation module. 
	The Meta learner is composed by the representation module and the AAO module, and the training process can be separated into two stages: pre-training stage and Meta training stage.
	
	In the pre-training stage, the representation module and the PC module will be trained on classification task on the MiniImagenet-900 dataset.
	
	After the pre-training stage, the representation module has learned knowledge from the MiniImagenet-900 dataset, and by utilizing the learned knowledge, the Meta learner can express the original input image into high representations which are suitable to differentiate different image classes. 
	In method RAML, the representation module can be built up by many kinds of network. We use the ResNet-50 network\cite{resnet} as the representation module in our paper.
	
	In the Meta training stage, for the Meta learner not forgetting the learned knowledge, we fix the pre-trained representation module totally, and the Meta learner only needs to learn how to solve the few shot learning task by quickly adjusting its AAO module based on the low dimensional meaningful features provided by the representation module, which is a simpler work compared with that of the Meta learner in AML method.
	
	It should be noted that the dataset used in the pre-training stage is different with that in the meta-training stage. In the meta training stage, the Meta learner is trained on the MiniImagenet dataset, whereas in the pre-training stage, representation module of the Meta learner is trained on MiniImagenet-900 dataset, and there is no image class overlaps between these two dataset. 
	
	\subsection{Attention model}
	\begin{figure}
		\centering
		\includegraphics[width=0.35\textwidth]{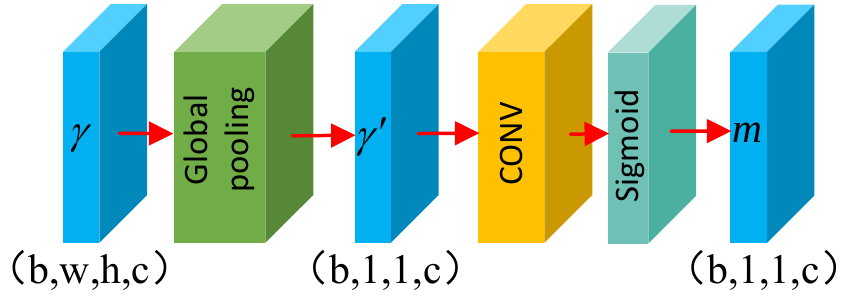}
		\caption{The network structure of attention model. The shape of $\gamma$ is (b,w,h,c) which is shown at the bottom of the figure, where b is the batch size, w and h are the size of feature map, c is the number of channels of the feature map, and the shape of $\gamma'$ and \emph{m} are both (b,1,1,c).}
		\label{fig:network structure of attention}
	\end{figure}
	In this paper, we use soft attention mechanism to build up the attention model, and the Figure\ref{fig:4b} is used to better understand the processing of the soft attention mechanism for Meta learner. Although the soft attention mechanism is not same with the attention mechanism in human vision system, it still plays similiar role with the human attention mechanism and will help the Meta learner to pay attention to the key features.
	The inner structure of attention model and the shape of corresponding features are shown in Figure.\ref{fig:network structure of attention}, and the computation process of attention model is shown as Eq.\ref{eq:attention model}. The feature $\gamma$ is firstly global-average-pooled to get feature $\gamma$$'$, and then a convolution layer coupled with a sigmoid activation layer are used to calculate the attention mask \emph{m} from the feature $\gamma$$'$. The attention model can be seen as a network that predict the importance of each channel of the feature map and get the attention mask \emph{m} by analyzing the total feature of the input data. 
	\begin{equation}
	\begin{split}
	\gamma' &= F_{avg}(\gamma), \\
	\emph{m} &= \sigma(F_{CNN}(\gamma'; \theta_{att})) 
	\label{eq:attention model}
	\end{split}
	\end{equation}
	
	Where \emph{$F_{avg}$} is the global-average-pooling function, and the $\sigma$ is the sigmoid activation function.
	
	Our attention model is different with SENet's\cite{soft_attention3}. In one aspect, our purpose of using attention model is to improve the Meta learner's ability to quickly adjust itself to pay attention to the important features of high representations, whereas SENet uses attention mechanism through all blocks. In another aspect, attention model's network structure is different, and we have found that SENet's attention model works not as good as ours in the few shot learning problem.
	
	\section{Experiments}
	In this section, we will present results and some details of our experiments, and the dataset we used. More details about our experiments will be provided in the supplementary material. All our experimental code is based on the Tensorflow library\cite{TensorFlow}. 
	\subsection{Dataset}
	We used several kinds of dataset in all our experiments: MiniImagenet\cite{miniimagenet}, Omniglot\cite{omniglot}, MiniImagenet-900 and Places2\cite{places365}, in all our experiments.
	
	\noindent\textbf{MiniImagenet}\cite{miniimagenet} is a dataset that popularly used for evaluating the performance of Meta learning algorithm, it contains totally 100 image classes, including 64 training classes, 12 validation classes, and 24 testing classes. Each image class with 600 images are sampled from the ImageNet dataset\cite{imagenet}.

	\noindent\textbf{Omniglot}\cite{omniglot} is another widely used dataset for few shot learning problem, it contains 50 different alphabets and totally 1623 characters from these alphabets, and each character has 20 images that hand drawn by 20 different people. 

	\noindent\textbf{MiniImagenet-900} dataset is designed to pre-train the representation module in method RAML, it is composed of 900 image classes. Each image class and the corresponding images is collected from the original ImageNet dataset, and each image class contains about 1300 images. It should be noted that there is no image class in MiniImageNet-900 is coincided with the classes from the MiniImagenet dataset. 
	
	\noindent\textbf{Places2}\cite{places365} is a dataset used for scene classification. In this paper, we also validate method RAML by pre-training the representation module on Place2 dataset. 
	
	In our work, for a fairly comparison with the previous few shot learning and Meta learning methods, we resize all the images of MiniImagenet, MiniImagenet-900 and Places2 to 84*84 resolution, and resize all the images of Omniglot to 28*28 resolution.
	
	\subsection{Experiment on MiniImagenet}
	On the MiniImagenet dataset, we test our method AML and RAML, and both of the two methods work very well and attain several state of the art performance on this dataset.
	
	\subsubsection{AML experiment}
	In the method AML, we improve the Meta learner's ability of learning from few data by explicitly embedding attention mechanism into its network, and train it by Meta-SGD approach. The structure of the Meta leaner's network is shown in Figure.\ref{fig:network structure of AML}, both the network of the attention mechanism and the classifier is a simple fully-connect layer. We set the hyper-parameter of K for 5way 1shot and 5way 5shot tasks to 1 and 5 respectively, whereas the L (image number of each class in query set) is always set to 15.
	
	The experimental result of our method AML on the MiniImagenet is shown in Tab.~\ref{tab:result on MiniImagenet}, we attained the state-of-the-art on the 5-way 5-shot image classification task: 69.46\%(compared with original Meta-SGD, we rise up the Meta learner's performance by 8.5\%). 
	
	\subsubsection{RAML experiment}
	In the method RAML, we give the Meta learner the ability of both utilizing the past knowledge and the attention mechanism by dividing the Meta learner's total network into representation module and AAO module, and these two modules are trained in different ways in two stages: pre-training stage and the Meta training stage. 
	
	In the pre-training stage of RAML method, the representation module and PC module were trained together by supervised learning with MiniImagenet-900 dataset. 
	The representation module is a modified ResNet-50 network which can be fed with the image of 84*84 resolution in RGB color space, and the PC module is a simple fully-connect layer followed by a Softmax-output layer. 
	In the Pre-training stage, we set the batch size to 256, and the learning rate to 0.001, and decay the learning rate to 0.0001 after 30000 iterations, and use L2 normalization and dropout operation to prevent over-fitting.
	Finally, the representation module together with the PC module get a 57.39\% classification accuracy on the 900 way classification task on the MiniImagenet-900 dataset. 
	
	After the pre-training stage, in the Meta training stage, we fix the weights of the representation module, and train the AAO module by Meta-SGD algorithm. The hyper-parameters of the Meta training process is shown in Tab.\ref{tab:Meta learning hyper parameters}. Experimental result is shown in Tab.\ref{tab:result on MiniImagenet}. Compared with method AML, method RAML improves the Meta learner's performance more greatly, the accuracy of 5-way 1-shot task rises up from 52.25\% to 63.66\%, and the accuracy of 5-way 5-shot task rises up from 69.46\% to 80.49\%.
	
	The most likely reason why RAML performs so well is: before the meta training stage, the representation module has learned the knowledge and the ability to understand the input image, and provides meaningful high representaions and features of the input image.
	In the Meta training stage, by fixing the representation module, the Meta learner's work becomes easier because it only needs to learn how to quickly adjust the AAO module according to the meaningful features the representation module provided, and do not need to take care of the original high dimensional input data. While the Meta learner trained by AML has to adjust the total network when faced with a new few shot learning task, and has to adjust its network and make decision according to the original high dimensional input data, which is a harder work than that of the Meta learner of RAML. 
	
	\begin{table}
		\centering
		\footnotesize
		\begin{tabular}{|c|c|c|}
			\hline
			\diagbox{Hyper parameters}{Dataset} & MiniImagenet & Omniglot \\
			\hline
			Meta learning rate & 0.001 & 0.001\\
			\hline
			total training tasks & 200000 & 200000\\
			\hline
			total val/test tasks & 600 & 600 \\
			\hline
			training iterators & 60000 & 40000 \\
			\hline
			L & 15 & 15 \\
			\hline
			L1 normalization & 0.001 & 0 \\
			\hline
			L2 normalization & 0.00001 & 0.00001 \\
			\hline
			Dropout & 0.2 & 0.2 \\
			\hline
		\end{tabular}
		\caption{Hyper parameters of the Meta learning process, which train the Meta learner on the MiniImagenet or Omniglot dataset with the AML or RAML methods}
		\label{tab:Meta learning hyper parameters}
	\end{table}
	
	\begin{table}
		\centering
		\resizebox{0.48\textwidth}{!}{
			\begin{tabular}{|c|c|c|c|}
				\hline
				\multirow{2}{*}{Method} &\multirow{2}{*}{Venue} &\multicolumn{2}{c|}{5-way Accuracy} \\
				\cline{3-4} & &1-shot &5-shot \\
				\hline
				matching nets FCE\cite{match-network}	&NIPS-16 &44.20\% &57.00\% \\
				\hline
				Meta learner LSTM\cite{miniimagenet}	&ICLR-17 &43.44$\pm$0.77\% &60.60 $\pm$0.71\% \\
				\hline
				MAML\cite{MAML} &ICML-17	             &48.70$\pm$1.84\% &63.11$\pm$0.92\% \\
				\hline
				Prototypical Nets\cite{prototypical}	&NIPS-17 &49.42$\pm$0.78\%	&68.20$\pm$0.66\% \\
				\hline
				Meta-SGD\cite{Meta-SGD} &        /         &50.47$\pm$1.87\%	&64.03$\pm$0.94\% \\
				\hline
				Reptile + Transduction\cite{Reptile} &  / &49.97$\pm$0.32\% &65.99$\pm$0.58\% \\
				\hline
				SNAIL\cite{SNAIL}& ICLR-18 &{\bfseries {\color{blue} 55.71$\pm$0.99\%}}  &68.88$\pm$0.92\% \\
				\hline
				RELATION NET\cite{comparenet} &CVPR-18	     &51.38$\pm$0.82\%  &67.07$\pm$0.69\% \\
				\hline
				GNN\cite{GNN} &ICLR-18	             &50.33$\pm$0.36\%	&66.41$\pm$0.63\% \\
				\hline
				AML(ours)&	/	         &52.25$\pm$0.85\%	&{\bfseries {\color{blue} 69.46$\pm$0.68\%} } \\
				\hline
				RAML(ours)	&/&{\bfseries 63.66$\pm$0.85\%}	&{\bfseries 80.49$\pm$0.45\%} \\
				\hline
			\end{tabular}
		}
		\caption{Few shot learning performance on MiniImagenet dataset. The accuracy is averaged by the accuracies on 600 few shot classification tasks, with 95\% confidence intervals, and all these 600 tasks are randomly generated from the test set of the MiniImagenet dataset. We highlight the best result and the second best result on each task.}
		\label{tab:result on MiniImagenet}
	\end{table}

	\begin{table*}
		\centering
		\resizebox{0.8\textwidth}{!}{
			\begin{tabular}{|c|c|c|c|c|c|}
				\hline
				\multirow{2}{*}{Method} &\multirow{2}{*}{Venue} &\multicolumn{2}{c|}{5-way Accuracy} &\multicolumn{2}{c|}{20-way Accuracy} \\
				\cline{3-6} & &1-shot &5-shot &1-shot &5-shot\\
				\hline
				matching nets FCE\cite{match-network}	&NIPS-16	&98.10\%	&98.90\%	&93.80\%	&98.50\% \\
				\hline
				MAML\cite{MAML}	&ICML-17	            &98.70$\pm$0.40\%	&99.90$\pm$0.10\%	&95.80$\pm$0.30\%	&98.90$\pm$0.20\% \\
				\hline
				Prototypical Nets\cite{prototypical}	&NIPS-17	&98.80\%	&99.70\%	&96.00\%	  &98.90\% \\
				\hline
				Meta-SGD\cite{Meta-SGD}	&/&99.53$\pm$0.26\%	&{\bfseries 99.93$\pm$0.09\%} &95.93$\pm$0.38\% &98.97$\pm$0.19\% \\
				\hline
				Reptile + Transduction\cite{Reptile}&/&97.68$\pm$0.04\%&99.48$\pm$0.06\%&89.43$\pm$0.14\%  &97.12$\pm$0.32\% \\
				\hline
				SNAIL\cite{SNAIL}&ICLR-18 &99.07$\pm$0.16\% &99.78$\pm$0.09\%&97.64$\pm$0.30\% &99.36$\pm$0.18\% \\
				\hline
				RELATION NET\cite{comparenet}	&CVPR-18	    &99.60$\pm$0.20\%	 &99.80$\pm$0.10\%   &97.60$\pm$0.20\%	&99.10$\pm$0.10\%  \\
				\hline
				GNN\cite{GNN}	&ICLR-18	            &99.20\%	&99.70\%	&97.40\%	&99.00\%  \\
				\hline
				AML(ours) &/	            &{\bfseries 99.65$\pm$0.10\%}	 &99.85$\pm$0.04\%	 &{\bfseries 98.48$\pm$0.09\%}  &{\bfseries 99.55$\pm$0.06\%} \\
				\hline
			\end{tabular}
		}
		\caption{Few shot learning performance on Omniglot dataset. The accuracy is tested by the same way in MAML\cite{MAML}}
		\label{tab:result on Omniglot}
	\end{table*}
	
	\subsection{Experiment on Omniglot}
	For the Omniglot dataset, we test AML on 5way 1shot, 5way 5shot, 20way 1shot, 20way 5shot tasks. By referencing to MAML\cite{MAML}, the architecture of the network we used in the Omniglot experiment is similar to that in the MiniImagenet dataset experiments. The hyper-parameter of the Meta training process is shown in Tab.\ref{tab:Meta learning hyper parameters}, the Meta batch size is set to 32 for 5way tasks, and 16 for 20way tasks. The experimental results is shown in Tab.\ref{tab:result on Omniglot}
	
	It is clear that in the 4 few shot image classifacation tasks, our method AML attain state-of-the-art performance on 3 of these 4 tasks. Especially on the 20-way 1-shot task, our method AML surpass other methods by a large margin (ompared with the result of original Meta-SGD, AML improves the Meta learner's performance from 95.93\% to 98.48\%). 
	
	\begin{figure*}
		\centering
		\includegraphics[width=0.95\textwidth]{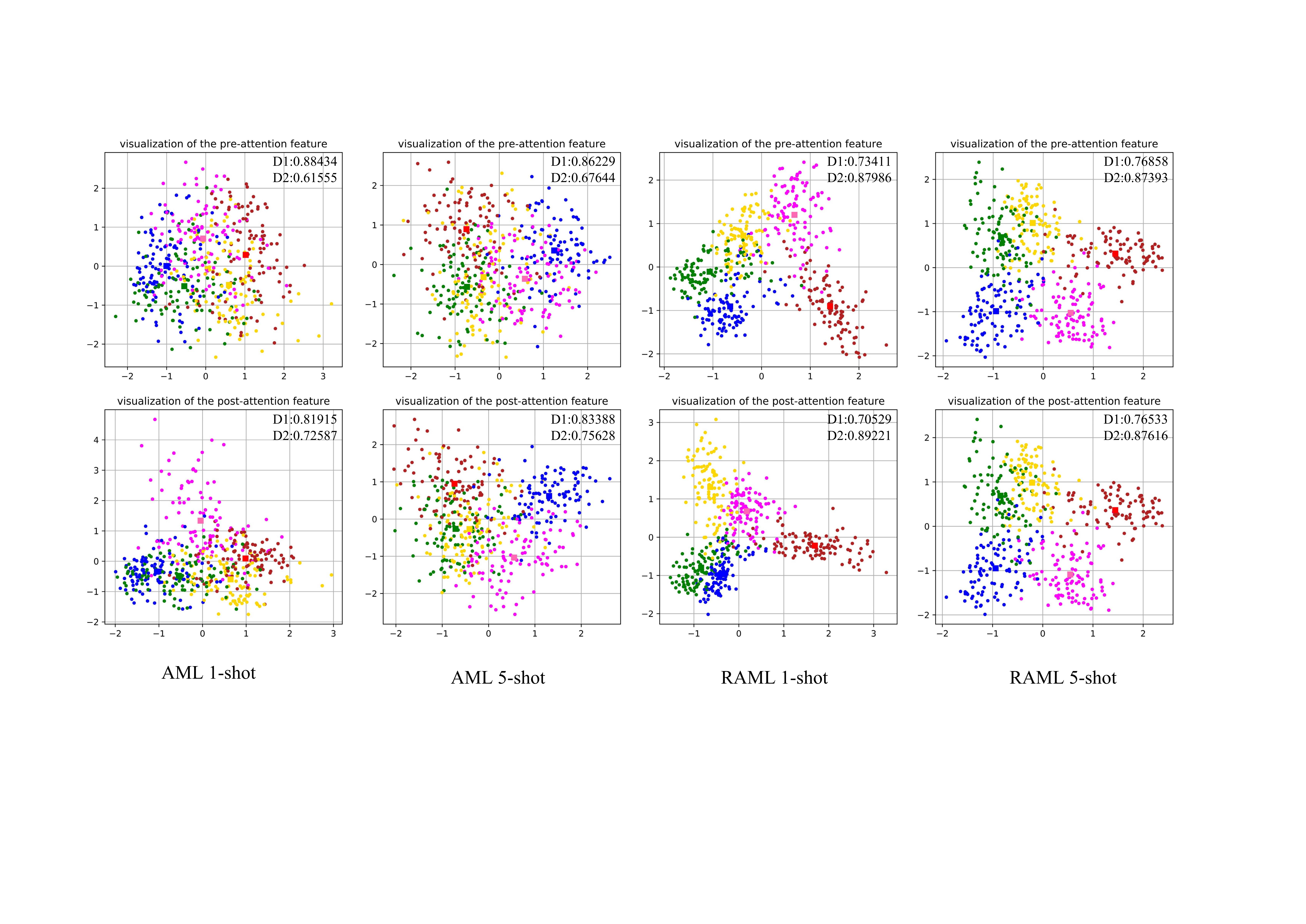}
		\caption{Visualization of the feature points of the pre-attention feature and the post attention feature of the AML and RAML experiments. The pre-attention feature is the feature $\gamma$ in Fig.\ref{fig:network structure of AML} and Fig.\ref{fig:network structure of RAML}, and the post-feature is the feature $\gamma_\alpha$. We use PCA algorithm to show the feature in a 2D space, and These feature points are colored by 5 colors, each color represents 1 image class of the 5 image class of the 5-way K-shot image classification task. D1 and D2 are standard deviation of the inner-class distance and the inter-class distance respectively. It is clear that the inner-class distance becomes smaller and the inter class distance becomes larger after the feature were operated by attention mechanism.}
		\label{fig:feature points}
	\end{figure*}

	\subsection{Ablation study}
	In this section, we will confirm the reliability of our methods by ablation experiments.
	\begin{table*}
		\centering
		\resizebox{0.7\textwidth}{!}{
			\begin{tabular}{|c|c|c|c|c|}
				\hline
				\multirow{2}{*}{Method} &\multicolumn{2}{c|}{5-way Accuracy} &\multicolumn{2}{c|}{20-way Accuracy} \\
				\cline{2-5} &1-shot &5-shot &1-shot &5-shot \\
				\hline
				MAML* &97.40$\pm$0.27\% &{\bfseries 99.71$\pm$0.05\%} &{\bfseries 93.37$\pm$0.23\%} &97.46$\pm$0.11\% \\
				\hline
				MAML+attention &{\bfseries 97.41$\pm$0.28\%} &99.48$\pm$0.12\% &92.99$\pm$0.25\% &{\bfseries 97.94$\pm$0.10\%} \\
				\hline
				Meta-SGD* &98.94$\pm$0.17\% &99.51$\pm$0.07\% &95.82$\pm$0.21\% &98.40$\pm$0.09\% \\
				\hline
				Meta-SGD+attention &{\bfseries 99.26$\pm$0.15\%} &{\bfseries 99.79$\pm$0.04\%} &{\bfseries 97.94$\pm$0.14\%} &{\bfseries 98.99$\pm$0.10\%} \\
				\hline
			\end{tabular}
		}
		\caption{The result of ablation experiments that test the effect of attention mechanism, on the Omniglot dataset}
		\label{tab:ablation result on Omniglot}
	\end{table*}
	
	\begin{table}
		\centering
		\resizebox{0.4\textwidth}{!}{
		\begin{tabular}{|c|c|c|}
			\hline
			\multirow{2}{*}{Method} &\multicolumn{2}{c|}{5-way Accuracy} \\
			\cline{2-3} &1-shot &5-shot \\
			\hline
			MAML* &48.03$\pm$0.83\% &64.11$\pm$0.73\% \\
			\hline
			MAML+attention &{\bfseries 48.52$\pm$0.85\%} &{\bfseries 64.94$\pm$0.69\%} \\
			\hline
			Reptile* &48.23$\pm$0.43\% &63.69$\pm$0.49\% \\
			\hline
			Reptile+attention &{\bfseries 48.30$\pm$0.45\%} &{\bfseries 64.22$\pm$0.39\%} \\
			\hline
			Meta-SGD* &48.15$\pm$0.93\% &63.73$\pm$0.85\% \\
			\hline
			Meta-SGD+attention &{\bfseries 49.11$\pm$0.94\%} &{\bfseries 65.54$\pm$0.84\%} \\
			\hline
		\end{tabular}
		}
		\caption{The results of ablation experiments that test the effect of attention mechanism, on the MiniImagenet dataset}
		\label{tab:ablation result on MiniImagenet}
	\end{table}
	
	\begin{table}
		\centering
		\resizebox{0.4\textwidth}{!}{
			\begin{tabular}{|c|c|c|}
				\hline
				\multirow{2}{*}{Method} &\multicolumn{2}{c|}{5-way Accuracy} \\
				\cline{2-3} &1-shot &5-shot \\
				\hline
				AML&{\bfseries 52.25$\pm$0.85\%}	&{\bfseries69.46$\pm$0.68\%} \\
				\hline
				AML(MAML)&50.65$\pm$0.92\%	&68.95$\pm$0.69\% \\
				\hline
				OML &51.27$\pm$0.78\%	&67.73$\pm$0.65\% \\
				\hline
				RAML	&63.66$\pm$0.85\%	&80.49$\pm$0.45\% \\
				\hline
				RAML(MAML)	&{\bfseries 64.23$\pm$0.85\%}	&{\bfseries83.76$\pm$0.49\%} \\
				\hline
				RAML(Places2)	&58.82$\pm$0.89\%	&74.09$\pm$0.76\% \\
				\hline
			\end{tabular}
		}
		\caption{Results of other ablation experiments}
		\label{tab:other ablation}
	\end{table}
	\subsubsection{Ablation study of AML}
	Firstly, to confirm the promotion effect of attention mechanism for the Meta learning algorithms, we do a lot of experiments to compare the performance of the Meta learner which is attention augmented with its counterpart which is not. The experimental results show in Tab.\ref{tab:ablation result on MiniImagenet} and Tab.\ref{tab:ablation result on Omniglot}. The performance of the method which has a * mark is the result re-implemented by ourselves. There are some difference between the result of the corresponding paper with that of our re-implementation, which is probably caused by different hyper-parameters or different experiment setting. The comparison results revealing that in most cases, the attention mechanism improves the Meta learner by a clear margin, demonstrating the reasonablility of our idea and method AML. 
	
	Furthermore, to test whether the method AML is universal to other Meta learning approaches, we change the Meta learning algorithm in AML from Meta-SGD to MAML, and all the network and hyper-parameters are constant. We mark it as AML(MAML). Corresponding result is shown in Tab.~\ref{tab:other ablation}. Though the performance of AML(MAML) drop down slightly, it is also comparable, which indicates that our method AML generalizes well to different Meta learning approaches. 
	
	As attention mechanism will bring more weights into the Meta learner's network, we do another experiment to validate that the improvement of AML is not caused by the growth of number of weights but the contribution of attention mechanism. The experiment detail is: since the attention model is mainly a convolution layer with kernel size of 1*1, we remove the attention model, and place a convolution layer with kernel size of 1*1 on the top of the feature $\gamma$(shown in Fig.\ref{fig:network structure of AML}). We name the Meta learner with this network as OML (Ordinary Meta learning), and its number of weight is the same with that of AML. Corresponding experimental result is shown in Tab.\ref{tab:other ablation}, and it is clear that OML lags behind AML, which shows that the improvement effect of AML is not caused by the growth of number of weight but the contribution of attention mechanism.
	
	\subsubsection{Ablation study of RAML}
	Similiarly to AML, we also test whether RAML can generalize well to other Meta learning approaches. We validated it by training the AAO module with MAML approach in the meta training stage. We mark it as RAML(MAML), and the experimental result is shown in Tab.\ref{tab:other ablation}.It is obviously that our method RAML also generalizes well to different Meta learning approaches.
	
	We do another experiment to test how the dataset that used in the pre-training stage affects the Meta learner. 
	We do this experiment by pre-training the representation module on the Places2\cite{places365} dataset, and all the other experiment settings and hyper-parameters are constant with primordial RAML, we mark it as RAML(Places2). Corresponding experimental result shows in Tab.\ref{tab:other ablation}. It is clear that the dataset used in the pre-training stage affects the Meta learner. 
	The possible reason is that different dataset used in the pre-training stage will lead the representation module to learn different knowledge and features of the input data\cite{Bolei2017Network}, and the places2 dataset is a dataset commonly used for scene classification, which result in the representation module to learn the knowledge about scene, and features which it outputs are more suitable for the scene classification task rather than the object classification task. 
	
	\subsection{Feature analyse}
	To understand the effect of attention mechanism, we reduce the feature $\gamma$ and $\gamma_\alpha$ (shown in Fig.\ref{fig:network structure of AML} and Fig.\ref{fig:network structure of RAML}) into a 2 dim space with PCA algorithm, and visualize them on a 2D plant. As shown in Fig.\ref{fig:feature points}, we visualize $\gamma$ and $\gamma_\alpha$ of the Meta learner trained on 5way 1 and 5 shot tasks with method AML and that with the method RAML, and each picture contains 500 feature points which represents 500 images of the query set. It is clear that the distribution of $\gamma_\alpha$ is more distinguishable between different image classes than $\gamma$, and the standard deviation of the inner-class distance becomes smaller and that of the inter-class distance becomes larger compared with that of $\gamma$. The reason of this phenomenon is simple: the attention mechanism makes the Meta learner pays more attention on the key feature, and the key feature will affect the $\gamma_\alpha$ more, which makes $\gamma_\alpha$ be more distinguishable than $\gamma$ to differentiate images of different classes.
	
	\section{Conclusion}
	In this paper, aiming to improve the computer vision system the ability of learning from few images, we analyze the problems of previous Meta learning approaches, and proposed our viewpoint that the attention mechanism is very helpful for Meta learner, besides, the Meta learner should be trained on the high representations of the input image instead of the original high dimensional RGB image, and it should own the ability of leveraging the past learned knowledge to accurately express the input image into high representaions. Based on our viewpoint, we design two methods: AML and RAML. Both of our methods work successful, and attain state-of-the-art performance on several few shot learning benchmarks, and revealing the reliability of our viewpoint and methods.

	\newpage
	
	{\small
		\bibliographystyle{ieee}
		\bibliography{main}

\begin{thebibliography}{10}\itemsep=-1pt

\bibitem{TensorFlow}
M.~Abadi, P.~Barham, J.~Chen, Z.~Chen, A.~Davis, J.~Dean, M.~Devin,
  S.~Ghemawat, G.~Irving, M.~Isard, et~al.
\newblock Tensorflow: a system for large-scale machine learning.
\newblock In {\em OSDI}, volume~16, pages 265--283, 2016.

\bibitem{Bolei2017Network}
D.~Bau, B.~Zhou, A.~Khosla, A.~Oliva, and A.~Torralba.
\newblock Network dissection: Quantifying interpretability of deep visual
  representations.
\newblock In {\em Computer Vision and Pattern Recognition}, pages 3319--3327,
  2017.

\bibitem{On-the-optimiazation}
S.~Bengio, Y.~Bengio, J.~Cloutier, and J.~Gecsei.
\newblock On the optimization of a synaptic learning rule.
\newblock In {\em Preprints Conf. Optimality in Artificial and Biological
  Neural Networks}, pages 6--8. Univ. of Texas, 1992.

\bibitem{Learning-a-synaptic}
Y.~Bengio, S.~Bengio, and J.~Cloutier.
\newblock {\em Learning a synaptic learning rule}.
\newblock Universit{\'e} de Montr{\'e}al, D{\'e}partement d'informatique et de
  recherche op{\'e}rationnelle, 1990.

\bibitem{seg2}
K.~J. Dai and Y.~L. R-FCN.
\newblock Object detection via region-based fully convolutional networks. arxiv
  preprint.
\newblock {\em arXiv preprint arXiv:1605.06409}, 2016.

\bibitem{imagenet}
J.~Deng, W.~Dong, R.~Socher, L.-J. Li, K.~Li, and L.~Fei-Fei.
\newblock Imagenet: A large-scale hierarchical image database.
\newblock In {\em Computer Vision and Pattern Recognition, 2009. CVPR 2009.
  IEEE Conference on}, pages 248--255. Ieee, 2009.

\bibitem{MAML}
C.~Finn, P.~Abbeel, and S.~Levine.
\newblock Model-agnostic meta-learning for fast adaptation of deep networks.
\newblock {\em arXiv preprint arXiv:1703.03400}, 2017.

\bibitem{GNN}
V.~Garcia and J.~Bruna.
\newblock Few-shot learning with graph neural networks.
\newblock {\em arXiv preprint arXiv:1711.04043}, 2017.

\bibitem{zhuang}
S.~Gidaris and N.~Komodakis.
\newblock Dynamic few-shot visual learning without forgetting.
\newblock In {\em Proceedings of the IEEE Conference on Computer Vision and
  Pattern Recognition}, pages 4367--4375, 2018.

\bibitem{resnet}
K.~He, X.~Zhang, S.~Ren, and J.~Sun.
\newblock Deep residual learning for image recognition.
\newblock In {\em Proceedings of the IEEE conference on computer vision and
  pattern recognition}, pages 770--778, 2016.

\bibitem{human_attention2}
S.~A. Hillyard, E.~K. Vogel, and S.~J. Luck.
\newblock Sensory gain control (amplification) as a mechanism of selective
  attention: electrophysiological and neuroimaging evidence.
\newblock {\em Philosophical Transactions of the Royal Society of London B:
  Biological Sciences}, 353(1373):1257--1270, 1998.

\bibitem{soft_attention3}
J.~Hu, L.~Shen, and G.~Sun.
\newblock Squeeze-and-excitation networks.
\newblock {\em arXiv preprint arXiv:1709.01507}, 7, 2017.

\bibitem{image-classification3}
G.~Huang, Z.~Liu, L.~Van Der~Maaten, and K.~Q. Weinberger.
\newblock Densely connected convolutional networks.
\newblock 1(2):3, 2017.

\bibitem{image-classification1}
A.~Krizhevsky, I.~Sutskever, and G.~E. Hinton.
\newblock Imagenet classification with deep convolutional neural networks.
\newblock In {\em Advances in neural information processing systems}, pages
  1097--1105, 2012.

\bibitem{omniglot}
B.~Lake, R.~Salakhutdinov, J.~Gross, and J.~Tenenbaum.
\newblock One shot learning of simple visual concepts.
\newblock In {\em Proceedings of the Annual Meeting of the Cognitive Science
  Society}, volume~33, 2011.

\bibitem{seg1}
Y.~Li, H.~Qi, J.~Dai, X.~Ji, and Y.~Wei.
\newblock Fully convolutional instance-aware semantic segmentation.
\newblock In {\em Computer Vision and Pattern Recognition}, pages 4438--4446,
  2017.

\bibitem{Meta-SGD}
Z.~Li, F.~Zhou, F.~Chen, and H.~Li.
\newblock Meta-sgd: Learning to learn quickly for few shot learning.
\newblock {\em arXiv preprint arXiv:1707.09835}, 2017.

\bibitem{human_attention1}
G.~Logan, D.~Dagenbach, and T.~Carr.
\newblock Inhibitory processes in attention, memory and language.
\newblock {\em Academic Press, San Diego}, pages 189--239, 1994.

\bibitem{SNAIL}
N.~Mishra, M.~Rohaninejad, X.~Chen, and P.~Abbeel.
\newblock A simple neural attentive meta-learner.
\newblock 2018.

\bibitem{hard_attention}
V.~Mnih, N.~Heess, A.~Graves, et~al.
\newblock Recurrent models of visual attention.
\newblock In {\em Advances in neural information processing systems}, pages
  2204--2212, 2014.

\bibitem{DRL1}
V.~Mnih, K.~Kavukcuoglu, D.~Silver, A.~Graves, I.~Antonoglou, D.~Wierstra, and
  M.~Riedmiller.
\newblock Playing atari with deep reinforcement learning.
\newblock {\em arXiv preprint arXiv:1312.5602}, 2013.

\bibitem{DRL2}
V.~Mnih, K.~Kavukcuoglu, D.~Silver, A.~A. Rusu, J.~Veness, M.~G. Bellemare,
  A.~Graves, M.~Riedmiller, A.~K. Fidjeland, G.~Ostrovski, et~al.
\newblock Human-level control through deep reinforcement learning.
\newblock {\em Nature}, 518(7540):529, 2015.

\bibitem{Reptile}
A.~Nichol, J.~Achiam, and J.~Schulman.
\newblock On first-order meta-learning algorithms.
\newblock 2018.

\bibitem{miniimagenet}
S.~Ravi and H.~Larochelle.
\newblock Optimization as a model for few-shot learning.
\newblock 2016.

\bibitem{An-alternative-to}
J.~Schmidhuber.
\newblock Learning to control fast-weight memories: An alternative to dynamic
  recurrent networks.
\newblock {\em Neural Computation}, 4(1):131--139, 1992.

\bibitem{prototypical}
J.~Snell, K.~Swersky, and R.~Zemel.
\newblock Prototypical networks for few-shot learning.
\newblock In {\em Advances in Neural Information Processing Systems}, pages
  4077--4087, 2017.

\bibitem{image-classification2}
C.~Szegedy, S.~Ioffe, V.~Vanhoucke, and A.~A. Alemi.
\newblock Inception-v4, inception-resnet and the impact of residual connections
  on learning.
\newblock In {\em AAAI}, volume~4, page~12, 2017.

\bibitem{attention-is-all-you-need}
A.~Vaswani, N.~Shazeer, N.~Parmar, J.~Uszkoreit, L.~Jones, A.~N. Gomez,
  {\L}.~Kaiser, and I.~Polosukhin.
\newblock Attention is all you need.
\newblock In {\em Advances in Neural Information Processing Systems}, pages
  5998--6008, 2017.

\bibitem{match-network}
O.~Vinyals, C.~Blundell, T.~Lillicrap, D.~Wierstra, et~al.
\newblock Matching networks for one shot learning.
\newblock In {\em Advances in Neural Information Processing Systems}, pages
  3630--3638, 2016.

\bibitem{soft_attention1}
F.~Wang, M.~Jiang, C.~Qian, S.~Yang, C.~Li, H.~Zhang, X.~Wang, and X.~Tang.
\newblock Residual attention network for image classification.
\newblock {\em arXiv preprint arXiv:1704.06904}, 2017.

\bibitem{face2}
F.~Wang, X.~Xiang, J.~Cheng, and A.~L. Yuille.
\newblock Normface: l 2 hypersphere embedding for face verification.
\newblock In {\em Proceedings of the 2017 ACM on Multimedia Conference}, pages
  1041--1049. ACM, 2017.

\bibitem{face1}
M.~Wang and W.~Deng.
\newblock Deep face recognition: A survey.
\newblock {\em arXiv preprint arXiv:1804.06655}, 2018.

\bibitem{soft_attention2}
T.~Xiao, Y.~Xu, K.~Yang, J.~Zhang, Y.~Peng, and Z.~Zhang.
\newblock The application of two-level attention models in deep convolutional
  neural network for fine-grained image classification.
\newblock In {\em Proceedings of the IEEE Conference on Computer Vision and
  Pattern Recognition}, pages 842--850, 2015.

\bibitem{comparenet}
F.~S.~Y. Yang, L.~Zhang, T.~Xiang, P.~H. Torr, and T.~M. Hospedales.
\newblock Learning to compare: Relation network for few-shot learning.
\newblock In {\em Proc. of the IEEE Conference on Computer Vision and Pattern
  Recognition (CVPR), Salt Lake City, UT, USA}, 2018.

\bibitem{places365}
B.~Zhou, A.~Lapedriza, A.~Khosla, A.~Oliva, and A.~Torralba.
\newblock Places: A 10 million image database for scene recognition.
\newblock {\em IEEE Transactions on Pattern Analysis and Machine Intelligence},
  2017.

\end{thebibliography}
	}
	
\end{document}